\documentclass[letterpaper, 10 pt, conference]{ieeeconf}  

\IEEEoverridecommandlockouts                              

\usepackage{cite}
\usepackage{graphics} 
\usepackage{epsfig} 
\usepackage{mathptmx} 
\usepackage{times} 
\usepackage{amsmath,amsfonts,amssymb}
\usepackage{booktabs}
\usepackage{multirow}
\usepackage{balance}
\usepackage{array}
\usepackage{stfloats}
\usepackage{url}
\usepackage{graphicx}

\title{\LARGE \bf
See Selectively, Act Adaptively: Dual-Level Structural Decomposition for Bimanual Robot Manipulation
}

\author{Yoon-Ji Choi, Young-Chae Son, Soo-Chul Lim*}

\begin{document}

\maketitle
\thispagestyle{empty}
\pagestyle{empty}

\begin{abstract}

In bimanual robotic manipulation, task-relevant visual information varies with the task stage and context, while the interaction of the two arms shifts between independent and coordinated modes, making policy learning challenging. However, existing monolithic VLA policies typically process multi-view observations and distinct bimanual interaction patterns without explicitly adapting their visual conditioning or action-generation structure. To address this issue, we propose a bimanual manipulation VLA framework based on Dual-Level Structural Decomposition. The View-Selective Visual Router dynamically adjusts wrist-view contributions to emphasize relevant visual cues, while the Interaction-Aware Action Mixture-of-Experts decomposes action generation into coordinated and arm-wise pathways to adapt to varying bimanual interaction modes. We evaluate the proposed method on six simulated bimanual manipulation tasks in RoboTwin 2.0 and three long-horizon real-world tasks. Our model improves the overall average success rate over a monolithic baseline by 27.7\% in simulation and 43.3\% in real-world evaluation, while consistently outperforming single-module variants across both settings. These results demonstrate that jointly considering selective visual processing and explicit decomposition of bimanual interaction structures provides an effective inductive bias for robust bimanual manipulation.

\end{abstract}

\section{Introduction}
\begin{figure}[!b]
\centering
\includegraphics[width=0.83\linewidth]{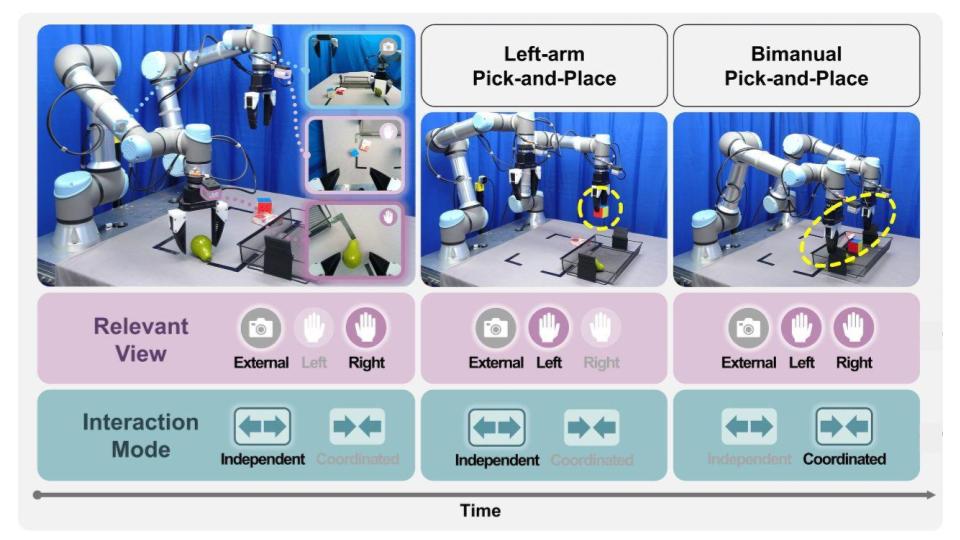}
\caption{\textbf{Motivation for dual-level structural decomposition.}
Task-relevant views and bimanual interaction modes vary across task stages.}
\label{fig:teaser}
\end{figure}

Bimanual robotic manipulation enables robots to perform independent and coordinated motions with two arms depending on the task stage and context~\cite{9849068,SMITH20121340}. Such capabilities are essential for complex manipulation tasks including object rearrangement, handover, and manipulation in which one arm acts while the other stabilizes an object~\cite{zhao2023learning,fu2024mobile}. Recent advances in imitation learning and VLA models have focused on learning visuomotor policies from human demonstrations, with extensions to bimanual settings~\cite{pmlr-v270-kim25c,pmlr-v305-black25a,liu2024rdt,chi2025diffusion}. However, bimanual manipulation is not merely a matter of doubling the degrees of freedom. When a task involves both arms, the required visual information changes across task stages, and spatial and temporal dependencies arise between the arms~\cite{10801861,pmlr-v305-deng25c,pmlr-v305-liu25f}. We characterize this challenge in terms of two forms of heterogeneity: perceptual heterogeneity and interaction heterogeneity, as illustrated in Fig.~\ref{fig:teaser}.

Perceptual heterogeneity refers to changes in the relevance of multi-camera inputs across task stages~\cite{11080054,Bai_2026_CVPR,chuang2025active}. For example, when one arm executes a grasp, the corresponding wrist view provides critical task cues, whereas the opposite wrist view may introduce task-irrelevant visual information. Existing monolithic policies often aggregate visual features without explicitly accounting for the time-varying relevance of each view. This can introduce task-irrelevant cues and weaken critical visual information, thereby degrading policy robustness and generalization~\cite{pmlr-v270-feng25a,son2026selectiveperceptionrobottaskaware}.

Interaction heterogeneity arises because bimanual tasks contain distinct action structures. Some stages allow the two arms to act independently, whereas others require coordinated motion toward a shared goal~\cite{9849068}. However, modeling both interaction modes within a shared action-generation pathway may lead to interference between mode-specific learning signals~\cite{zhai2026skillvlatacklingcombinatorialdiversity,Jiang_2025_ICCV,pmlr-v270-lee25a,xu2026movethenoperatebehavioralphasinghumanlike}.

In bimanual manipulation, these two forms of heterogeneity are intertwined~\cite{xiong2025vision,chuang2025active,pmlr-v270-liu25i}. For example, when one arm stabilizes an object while the other manipulates it, the actions of the two arms become more interdependent, and wrist-view cues from a specific arm can become critical. To address this issue, we propose a Dual-Level Structural Decomposition Framework that introduces structural specialization at both the perceptual and action levels. We introduce the View-Selective Visual Router (VSR) at the perceptual level, and the Interaction-Aware Action Mixture-of-Experts (IAMoE) at the action level. Together, they mitigate the influence of task-irrelevant visual cues and reduce interference among distinct interaction patterns. By jointly accounting for changing visual cues and evolving interaction structures, these two modules enable more robust policy learning in complex bimanual tasks. The main contributions of this paper are as follows:

\begin{itemize}
    \item
    We propose Dual-Level Structural Decomposition, combining task-aware visual routing with interaction-aware action decomposition in a bimanual VLA.

    \item
    We show that the perceptual- and action-level decompositions provide complementary benefits and consistently outperform monolithic and single-module variants.

    \item
    We validate the framework on six simulated and three long-horizon real-world tasks, demonstrating substantially improved performance under challenging visual variations.
\end{itemize}

\section{Related Work}

\subsection{Bimanual Policy Learning and Generalist VLAs}

Early bimanual imitation-learning approaches such as ACT~\cite{zhao2023learning} demonstrated that coordinated dual-arm behaviors can be learned directly from human demonstrations using a unified action-chunking policy. More recent approaches have scaled bimanual policy learning through larger models and broader datasets. RDT~\cite{liu2024rdt}, for example, develops a diffusion foundation model specifically for bimanual manipulation, while generalist VLA models such as OpenVLA~\cite{pmlr-v270-kim25c} and $\pi_{0.5}$~\cite{pmlr-v305-black25a} aim to learn visuomotor behaviors across broader task distributions and robot embodiments.

Another line of work improves bimanual learning by reusing single-arm knowledge. TwinVLA~\cite{im2026twinvladataefficientbimanualmanipulation} constructs bimanual behavior from two single-arm VLA policies, while AnyBimanual~\cite{Lu_2025_ICCV} transfers unimanual policies to bimanual settings. SkillVLA~\cite{zhai2026skillvlatacklingcombinatorialdiversity} further addresses the combinatorial diversity of dual-arm manipulation through skill reuse. These approaches primarily focus on data efficiency, transfer, or generalization, whereas our work focuses on structural variation within a task as the two arms alternate between independent and coordinated interaction.

\subsection{Selective Perception for Multi-View Manipulation}

Multi-view observations provide complementary global and local information for robot manipulation, but not all views are equally informative throughout task execution. BFA~\cite{11080054} performs feature-aware fusion to emphasize informative views for fine-grained manipulation, while stage-guided fusion methods~\cite{pmlr-v270-feng25a} adapt sensory contributions according to the current manipulation stage. Selective Perception~\cite{son2026selectiveperceptionrobottaskaware} explicitly routes task-relevant sensory information for multimodal VLA policies. Active-vision approaches further improve observation quality by adapting the viewpoint itself~\cite{Bai_2026_CVPR,chuang2025active}. These studies establish the benefit of adapting perception to task-dependent visual relevance. However, they primarily treat view selection or fusion as a perceptual problem, without explicitly considering its relationship with the downstream bimanual interaction structure. Our work instead considers task-dependent visual relevance together with interaction-dependent action generation.

\subsection{Structured Bimanual Interaction Modeling}

Several studies explicitly incorporate inter-arm structure into bimanual policy learning. Stabilize to Act~\cite{grannen2023stabilize} exploits asymmetric acting and stabilizing roles to coordinate the two arms, and VoxAct-B~\cite{pmlr-v270-liu25i} similarly models bimanual manipulation through structured functional roles. DIF~\cite{Jiang_2025_ICCV} introduces a decoupled interaction framework for bimanual manipulation, while InterACT~\cite{pmlr-v270-lee25a} models inter-arm dependency through hierarchical attention and action chunking. Move-Then-Operate~\cite{xu2026movethenoperatebehavioralphasinghumanlike} introduces behavioral phasing to separate different stages of manipulation. These approaches demonstrate that explicitly modeling bimanual structure can improve over fully monolithic action generation. However, many rely on fixed arm-wise decompositions, predefined roles, or predefined behavioral phases. In contrast, our framework focuses on tasks in which the required interaction structure changes during execution, and adapts action generation between independent and coordinated modes accordingly.

Taken together, prior work has studied selective perception and structured bimanual policy learning largely as separate problems. Our Dual-Level Structural Decomposition jointly addresses visual conditioning and action generation, coupling task-dependent view relevance with interaction-dependent bimanual structure within a single VLA framework.

\begin{figure*}[t]
    \centering
    \includegraphics[width=0.86\textwidth]
    {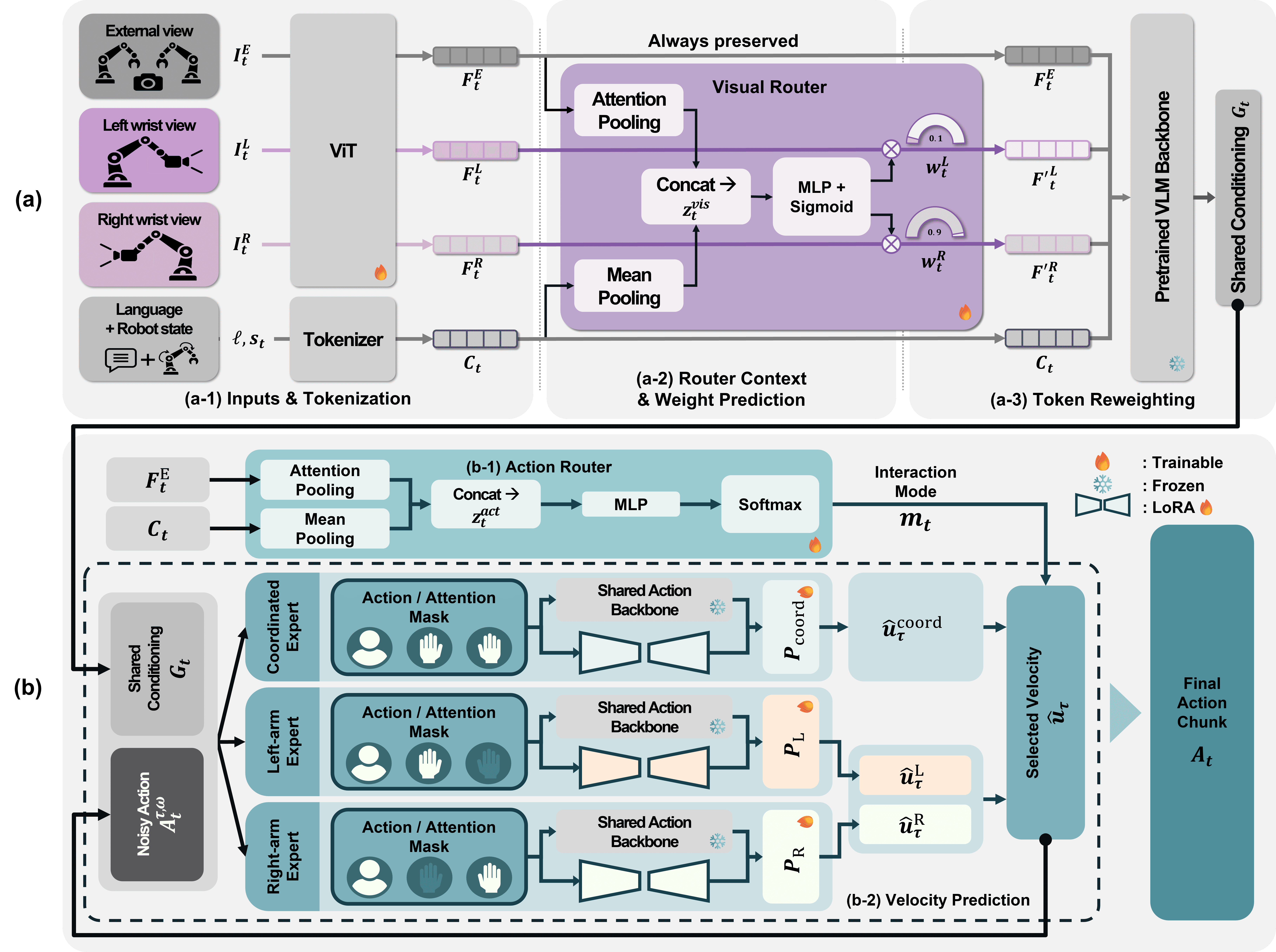}
    \caption{\textbf{Architecture of the proposed Dual-Level Structural Decomposition framework.}
    (a) The View-Selective Visual Router (VSR) predicts relevance weights for both wrist views and reweights their tokens before VLM conditioning.
    (b) The Interaction-Aware Action Mixture-of-Experts (IAMoE) routes action generation to coordinated or arm-wise experts.
    Expert-specific view attention and action masks couple interaction structure with visual conditioning: the coordinated expert accesses the full bimanual context, whereas each arm-wise expert operates on its corresponding view and action subspaces.}
    \label{fig:method_architecture}
\end{figure*}
\section{Method}
\label{sec:method}
In this section, we present Dual-Level Structural Decomposition, which introduces structural specialization at both the perceptual and action levels of a pretrained VLA. As illustrated in Fig.~\ref{fig:method_architecture}, the framework jointly adapts visual conditioning and bimanual action generation while preserving view-associated visual structure across the two levels.

\subsection{Problem Formulation and Architecture}

Given the view set \(V=\{\mathrm{E},\mathrm{L},\mathrm{R}\}\) (external, left, right), the observation at time \(t\) is given by \(O_t = (\{I_t^v\}_{v \in V},\, \ell,\, s_t)\). \(I_t^v\), \(\ell\), and \(s_t\) denote the image from view \(v\), the language instruction, and the robot state, respectively. From \(O_t\), the proposed model predicts an \(H\)-step action chunk \(A_t \equiv [a_t, a_{t+1}, \ldots, a_{t+H-1}]\), including joint control commands and gripper states for both arms. Given a demonstration dataset \(\mathcal{D}=\{(O_t,A_t)\}_{t=1}^{N}\), our goal is to learn a policy that accounts for view relevance and the bimanual interaction structure, both of which vary with the task context.

We adopt the pretrained \(\pi_{0.5}\)~\cite{pmlr-v305-black25a} as the base policy and introduce VSR and IAMoE as lightweight structural adaptations, with the aim of preserving the pretrained representation. Specifically, VSR modifies the visual-conditioning stage by modulating wrist-view contributions according to their task-dependent relevance, while IAMoE extends the action-generation stage with coordinated and arm-wise pathways.

Given the external, left-wrist, and right-wrist visual tokens \(F_t^{\mathrm{E}}, F_t^{\mathrm{L}}, F_t^{\mathrm{R}}\), and the language-state tokens \(C_t\), VSR reweights the wrist-view tokens before VLM conditioning (Fig.~\ref{fig:method_architecture}(a)). The view-associated token structure is retained through the VLM backbone, allowing each IAMoE pathway to operate with an interaction-dependent visual conditioning scope.

\subsection{View-Selective Visual Router (VSR)}
\label{sec:vsr}

The relevance of wrist observations changes throughout task execution. VSR therefore assigns an independent relevance weight to each wrist view, while retaining the external view as a global observation of the workspace.

VSR uses the external visual tokens \(F_t^{\mathrm{E}}\) and language-state tokens \(C_t\) as inputs (Fig.~\ref{fig:method_architecture}(a-2)). The external tokens are summarized using attention pooling, while the language-state tokens are summarized using mean pooling.
The resulting representations form the routing context \(z_t^{\mathrm{vis}}\) for VSR. The wrist-view tokens are not used as routing inputs. From this context, VSR predicts the wrist-view relevance weights.
\begin{equation}
    w_t =
    [w_t^{\mathrm{L}}, w_t^{\mathrm{R}}]
    \in [0,1]^2
\end{equation}
The weights are obtained using an MLP followed by an element-wise sigmoid. Because the relevance of the two wrist views is not mutually exclusive, both views can be simultaneously emphasized or attenuated.
\begin{equation}
\begin{aligned}
    F_t^{\prime\mathrm{L}} &= w_t^{\mathrm{L}} F_t^{\mathrm{L}}, &
    F_t^{\prime\mathrm{R}} &= w_t^{\mathrm{R}} F_t^{\mathrm{R}}.
\end{aligned}
\label{eq:vsr}
\end{equation}

The external tokens \(F_t^{\mathrm{E}}\), reweighted wrist tokens \(F_t^{\prime\mathrm{L}},F_t^{\prime\mathrm{R}}\), and language-state tokens \(C_t\) are processed by the pretrained VLM to form the conditioning tokens \(G_t\) (Fig.~\ref{fig:method_architecture}(a-3)).
Although these tokens are jointly processed by the VLM, their view-associated positions are retained. This allows IAMoE to impose expert-specific visual conditioning scopes at the subsequent action-generation stage.

\subsection{Interaction-Aware Action Mixture-of-Experts (IAMoE)}
\label{sec:iamoe}
IAMoE specializes action generation according to whether the two arms should act independently or as a coordinated system. Following the conditional flow-matching formulation of the base policy~\cite{0f001e52f7724ff3a539cdef54b4a160,tong2024improving}, the action expert predicts a target velocity field $u_\tau$ for the noisy action chunk $A_t^{\tau,\omega}$.

IAMoE contains one coordinated expert and two arm-wise experts~\cite{shazeer2017} (Fig.~\ref{fig:method_architecture}(b-2)). All three branches share the action backbone but use separate LoRA adapters~\cite{hu2022lora,wu2024mixture}. The coordinated expert receives the full visual conditioning context, including the external and both wrist views. In contrast, the left and right arm-wise experts attend to the external view and the wrist-view tokens associated with their respective arms, while the opposite wrist view is masked. Thus, the action pathway selected by IAMoE also determines the visual conditioning scope used for action generation.

The experts are similarly separated in action space. Let $M^{\mathrm{L}}, M^{\mathrm{R}} \in \{0,1\}^{H \times d}$ denote fixed masks selecting the left-arm and right-arm action subspaces. The coordinated expert operates on the complete bimanual action space, whereas each arm-wise expert operates only on its corresponding subspace.
Their velocity predictions are
\begin{equation}
\begin{gathered}
    \hat{u}_{\tau}^{\mathrm{coord}}
    = P_{\mathrm{coord}}(h_t^{\mathrm{coord}}),\\
    \hat{u}_{\tau}^{\mathrm{L}}
    = M^{\mathrm{L}} \odot P_{\mathrm{L}}(h_t^{\mathrm{L}}),\\
    \hat{u}_{\tau}^{\mathrm{R}}
    = M^{\mathrm{R}} \odot P_{\mathrm{R}}(h_t^{\mathrm{R}}),\\
    \hat{u}_{\tau}^{\mathrm{ind}}
    = \hat{u}_{\tau}^{\mathrm{L}} + \hat{u}_{\tau}^{\mathrm{R}}.
\end{gathered}
\label{eq:experts}
\end{equation}

The view and action masks are fixed structural masks. In our implementation, the 14-D bimanual action space is partitioned into two 7-D arm--gripper subspaces for the arm-wise experts, whereas the coordinated expert operates on the full 14-D action space.

To select between the coordinated and independent action structures, the Action Router predicts a hard one-hot interaction mode from the routing context $z_t^{\mathrm{act}}$ (Fig.~\ref{fig:method_architecture}(b-1)). The router uses the same contextual input sources as VSR but has independent parameters. The predicted mode is
\begin{equation}
    m_t =
    [m_t^{\mathrm{coord}}, m_t^{\mathrm{ind}}]
    \in \{[1,0], [0,1]\}.
\end{equation}
Accordingly, the final velocity prediction is given by
\begin{equation}
    \hat{u}_\tau =
    m_t^{\mathrm{coord}}\hat{u}_\tau^{\mathrm{coord}}
    +
    m_t^{\mathrm{ind}}\hat{u}_\tau^{\mathrm{ind}}.
    \label{eq:action_routing}
\end{equation}

During training, we use straight-through Gumbel-Softmax to preserve hard routing in the forward pass while allowing gradients to propagate through a soft relaxation~\cite{jang2017categorical}. At inference time, the interaction mode is selected by $\arg\max$ and held fixed throughout the current action chunk.

\subsection{Router Supervision and Training}
\label{sec:training}

The two routers are trained with explicit supervision corresponding to the two forms of structure targeted by our framework. The view-relevance label \(y_t^{\mathrm{vis}}\in\{0,1\}^2\) indicates whether each wrist view provides visual information required for the current or subsequent action. A wrist view is labeled relevant when it contains local cues needed for object interaction, approach, grasping, or target-directed motion. When one view provides a clearer observation of an occluded target or contact region, that view is labeled relevant regardless of which arm performs the action. If the two wrist views provide complementary information, both are labeled relevant; if their information is redundant, only the more informative view is retained. The interaction-mode label \(y_t^{\mathrm{act}}\in\{[1,0],[0,1]\}\) indicates whether the two arms can be controlled independently or must be considered jointly. A segment is labeled independent when the actions of the two arms can be generated separately without explicitly modeling inter-arm action coupling, such as when the arms move toward different objects or target locations. A segment is labeled coordinated when the actions of the two arms must be generated jointly to satisfy a shared manipulation objective or inter-arm constraint, such as joint object manipulation, stabilization, or handover.

Manually annotating these labels at every timestep is costly. We therefore generate router supervision offline using task-specific KNN label propagation from a small set of manually annotated reference episodes. For each task, \(10\%\) of the demonstration episodes are annotated at the frame level and used as the reference set. Separate KNN classifiers are constructed for the view-relevance and interaction-mode labels. At each timestep, we form a proprioceptive feature vector from the gripper state and velocity and the end-effector position and velocity of both arms, together with neighboring timesteps within a temporal context window. Although view relevance is defined from visual evidence, proprioception is used only as a proxy for matching corresponding task phases across demonstrations. Within a given task, changes in view relevance tend to align consistently with recurring arm states and motion patterns, allowing labels from visually annotated reference episodes to be propagated to demonstrations with similar proprioceptive trajectories.

We use a two-pass propagation procedure to improve label consistency near stage transitions. The first-pass KNN predicts labels over the full episode, after which temporal majority-vote smoothing is applied to suppress isolated prediction noise. We then identify label-transition frames from the first-pass sequence and define a local transition region around each boundary. A second KNN is trained with increased emphasis on transition samples and applied only within these regions, using a wider temporal context and a smaller neighborhood to improve sensitivity to boundary changes. The second-pass predictions replace the first-pass labels inside the transition regions, followed by a final local smoothing step. The sensitivity of this procedure to the amount of manual supervision and its agreement with manual labels during closed-loop rollouts are evaluated in Sec.~\ref{sec:experiments}.

The training objective consists of the router-conditioned action loss, the  router-supervision losses, and an auxiliary branch loss. We denote the standard conditional flow-matching loss of the base policy by \(\mathcal{L}_{\mathrm{main}}\). For VSR, binary cross-entropy is applied independently to the two view-relevance labels, yielding \(\mathcal{L}_{\mathrm{vis}}\). For the Action Router, cross-entropy over the interaction mode gives \(\mathcal{L}_{\mathrm{act}}\).

To prevent under-training of individual experts due to unstable routing early in training, we additionally supervise all expert branches, with the auxiliary supervision gradually reduced as the routing policy stabilizes:
\begin{equation}
\begin{gathered}
\mathcal{L}_{\mathrm{aux}}
=
\mathbb{E}\Big[
\|\hat{u}_{\tau}^{\mathrm{coord}}-u_{\tau}\|_2^2\\
+\|\hat{u}_{\tau}^{\mathrm{L}}
-M^{\mathrm{L}}\odot u_{\tau}\|_2^2
+\|\hat{u}_{\tau}^{\mathrm{R}}
-M^{\mathrm{R}}\odot u_{\tau}\|_2^2
\Big].
\end{gathered}
\label{eq:aux_loss}
\end{equation}

The total objective is
\begin{equation}
\mathcal{L}_{\mathrm{total}}
=
\mathcal{L}_{\mathrm{main}}
+
\lambda_{\mathrm{vis}}\mathcal{L}_{\mathrm{vis}}
+
\lambda_{\mathrm{act}}\mathcal{L}_{\mathrm{act}}
+
\lambda_{\mathrm{aux}}\mathcal{L}_{\mathrm{aux}},
\label{eq:total_loss}
\end{equation}
where \(\lambda_{\mathrm{vis}}\), \(\lambda_{\mathrm{act}}\), and \(\lambda_{\mathrm{aux}}\) control the corresponding objectives. The weight \(\lambda_{\mathrm{aux}}\) is gradually decreased during training so that the final optimization increasingly relies on the router-selected prediction pathway.
\begin{figure}[!b]
    \centering
    \includegraphics[width=0.95\linewidth,trim=0 42bp 0 42bp,clip]{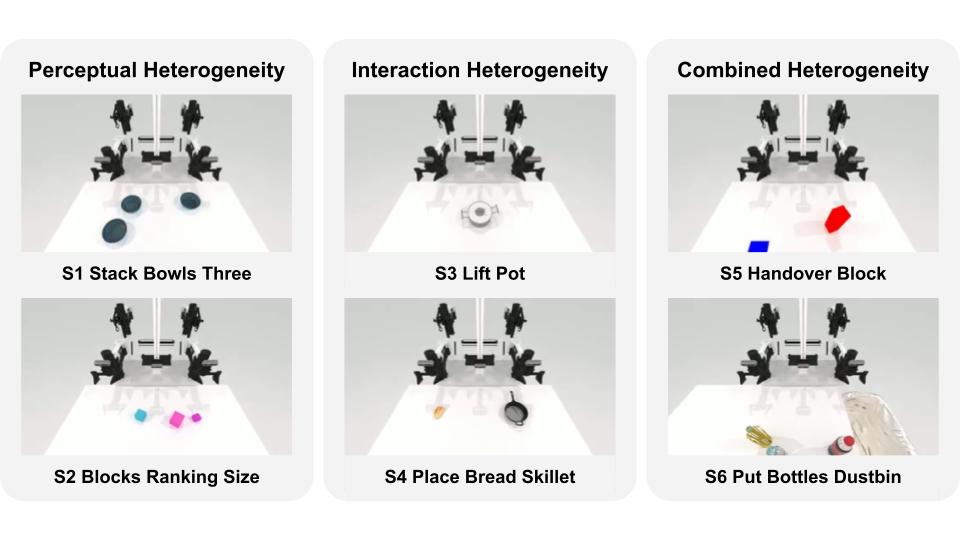}
    \caption{\textbf{Simulation tasks from RoboTwin~2.0.}}
    \label{fig:sim_tasks}
\end{figure}

\section{Experiments}
\label{sec:experiments}

We evaluate the proposed framework in simulation and on a real bimanual robot. The experiments are designed to answer four questions:
(1) Does Dual-Level Structural Decomposition improve bimanual manipulation performance over existing policies and a matched monolithic VLA baseline?
(2) What aspects of the proposed structure account for these gains, and are both decomposition levels and their coupling necessary?
(3) Does the proposed framework remain effective under challenging visual variations?
(4) How much manual supervision is required for the routers, and how reliably do they operate during closed-loop execution?

\subsection{Experimental Setup}

\textbf{Simulation tasks.}
We use six bimanual manipulation tasks from RoboTwin~2.0~\cite{chen2025robotwin}. The tasks are selected as a diagnostic subset covering different forms of heterogeneity: S1 (\emph{Stack Bowls Three}) and S2 (\emph{Blocks Ranking Size}) primarily exhibit perceptual heterogeneity; S3 (\emph{Lift Pot}) and S4 (\emph{Place Bread Skillet}) primarily require changes in inter-arm coordination; and S5 (\emph{Handover Block}) and S6 (\emph{Put Bottles Dustbin}) contain both forms, as illustrated in Fig.~\ref{fig:sim_tasks}. All policies are trained on demonstrations from the Easy setting and evaluated under both the Easy and Hard protocols.
\begin{figure*}[!t]
\centering
\includegraphics[width=0.84\textwidth]{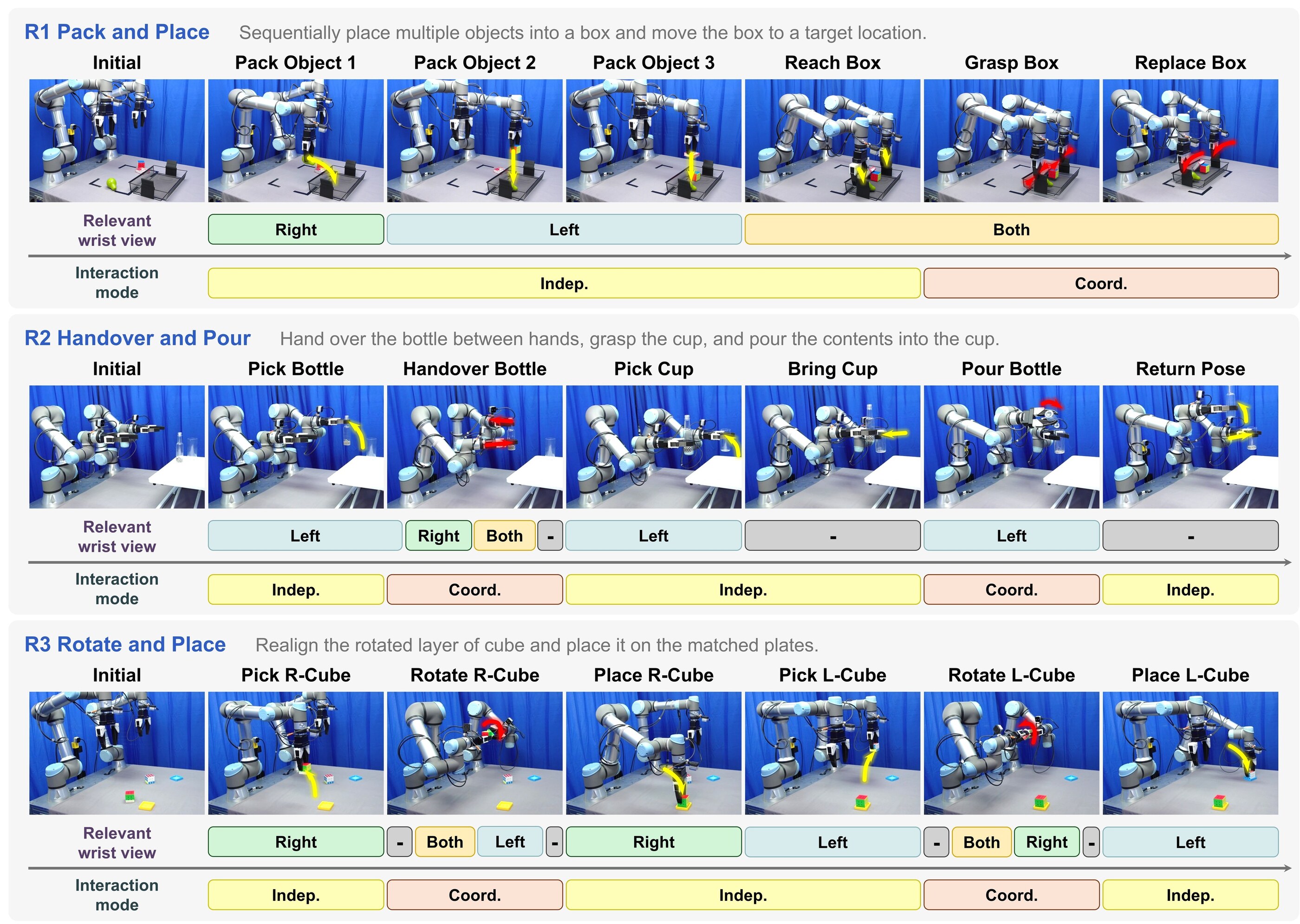}
\caption{\textbf{Real-world tasks R1--R3.}
For each illustrated execution, bars indicate the annotated wrist-view relevance and interaction mode.}
\label{fig:real}
\end{figure*}

\textbf{Real-world tasks.}
The real-world evaluation consists of three six-stage long-horizon tasks: R1 (\emph{Pack and Place}), R2 (\emph{Handover and Pour}), and R3 (\emph{Rotate and Place}), shown in Fig.~\ref{fig:real}. The relevant wrist view and interaction mode change repeatedly during each rollout. Object positions are randomized across episodes, which can alter both the execution order and the arm responsible for a particular motion. R1 and R3 use an \(800\,\mathrm{mm}\times400\,\mathrm{mm}\) workspace, while R2 uses an \(800\,\mathrm{mm}\times200\,\mathrm{mm}\) workspace with the bottle initialized in front of the cup. The Hard setting additionally replaces the training objects with unseen instances and introduces four distractor objects into each workspace. 

\textbf{Baselines and ablations.}
Table~\ref{tab:sim} compares our method with two groups of methods. The first group consists of representative prior methods---ACT~\cite{zhao2023learning}, Diffusion Policy (DP)~\cite{chi2025diffusion}, RDT~\cite{liu2024rdt}, TwinVLA~\cite{im2026twinvladataefficientbimanualmanipulation}, InterACT~\cite{pmlr-v270-lee25a}, and DIF~\cite{Jiang_2025_ICCV}---trained and evaluated using the same demonstration data and rollout protocol. The second group consists of controlled variants sharing the same
\(\pi_{0.5}\) backbone~\cite{pmlr-v305-black25a}: \emph{Baseline} is the monolithic fine-tuned policy; \emph{Ours w/o IAMoE} uses VSR alone; \emph{Ours w/o VSR} uses IAMoE alone; and \emph{Ours} uses both modules. \emph{LoRA \(2\times\)} increases the LoRA capacity of the monolithic baseline and slightly exceeds Ours in trainable parameter count, testing whether the gains are explained by additional capacity. \emph{Ours w/o view masks} retains both routers and action experts but removes the expert-specific view masks, directly testing the structural coupling between visual conditioning and action generation.

\textbf{Training and evaluation protocol.}
All methods are trained separately for each simulation task using the same 50 demonstrations, while the real-world variants use 40 demonstrations per task. The \(\pi_{0.5}\)-based variants take three \(224\times224\) images, use an action-chunk length of \(H=50\), and are fine-tuned with AdamW for 30,000 steps using a batch size of 16. Simulation evaluation uses 100 rollouts per task, while real-world evaluation uses 10 trials per task.

\begin{table*}[!t]
\centering
\caption{Simulation success rates (\%) on six RoboTwin~2.0 tasks. \(\Delta_{\mathrm{base}}\) denotes the average change from Baseline (\%).}
\label{tab:sim}
\footnotesize
\renewcommand{\arraystretch}{1.00}
\setlength{\tabcolsep}{1.1pt}
\begin{tabular*}{\textwidth}{@{\extracolsep{\fill}}l*{8}{c}@{\hspace{5pt}}*{8}{c}@{}}
\toprule
& \multicolumn{8}{c}{\textbf{Easy setting}} & \multicolumn{8}{c}{\textbf{Hard setting}} \\
\cmidrule(lr){2-9} \cmidrule(l){10-17}
Method & S1 & S2 & S3 & S4 & S5 & S6 & Avg. & $\Delta_{\mathrm{base}}$
& S1 & S2 & S3 & S4 & S5 & S6 & Avg. & $\Delta_{\mathrm{base}}$ \\
\midrule
ACT~\cite{zhao2023learning} 
& 48 & 0 & 88 & 7 & 42 & 27 & 35.3 & -- 
& 0 & 0 & 0 & 0 & 0 & 1 & 0.2 & -- \\

DP~\cite{chi2025diffusion} 
& 63 & 1 & 39 & 11 & 10 & 22 & 24.3 & -- 
& 0 & 0 & 0 & 0 & 0 & 0 & 0.0 & -- \\

RDT~\cite{liu2024rdt} 
& 51 & 0 & 72 & 5 & 45 & 21 & 32.3 & -- 
& 17 & 0 & 9 & 1 & 14 & 4 & 7.5 & -- \\

TwinVLA~\cite{im2026twinvladataefficientbimanualmanipulation} 
& 77 & 3 & 87 & 20 & 17 & 8 & 35.3 & -- 
& 15 & 0 & 7 & 1 & 0 & 4 & 4.5 & -- \\

InterACT~\cite{pmlr-v270-lee25a} 
& 84 & 0 & 91 & 5 & 49 & 56 & 47.5 & -- 
& 0 & 0 & 21 & 0 & 1 & 6 & 4.7 & -- \\

DIF~\cite{Jiang_2025_ICCV} 
& 69 & 4 & 96 & 55 & 54 & 70 & 58.0 & -- 
& 2 & 0 & 0 & 0 & 0 & 3 & 0.8 & -- \\

\midrule

Baseline ($\pi_{0.5}$)~\cite{pmlr-v305-black25a} 
& 67 & 27 & 97 & 63 & 69 & 46 & 61.5 & -- 
& 36 & 10 & 47 & 8 & 9 & 24 & 22.3 & -- \\

LoRA $2\times$ 
& 67 & 27 & 98 & 65 & 66 & 47 & 61.7 & +0.2 
& 35 & 9 & 47 & 6 & 8 & 21 & 21.0 & -1.3 \\

Ours w/o IAMoE 
& 71 & 50 & 95 & 65 & 75 & 51 & 67.8 & +6.3 
& 45 & 47 & 70 & 29 & 21 & 32 & 40.7 & +18.4 \\

Ours w/o VSR 
& 78 & 49 & \textbf{99} & 70 & 86 & 65 & 74.5 & +13.0 
& 50 & 33 & 75 & 38 & 33 & 40 & 44.8 & +22.5 \\

Ours w/o view masks
& 80 & 51 & 97 & 68 & 80 & 61 & 72.8 & +11.3
& 53 & 49 & 68 & 33 & 22 & 29 & 42.3 & +20.0 \\

\textbf{Ours} 
& \textbf{85} & \textbf{63} & 97 & \textbf{78} & \textbf{93} & \textbf{71} & \textbf{81.2} & \textbf{+19.7}
& \textbf{68} & \textbf{55} & \textbf{83} & \textbf{47} & \textbf{46} & \textbf{49} & \textbf{58.0} & \textbf{+35.7} \\

\bottomrule
\end{tabular*}
\vspace{-3pt}
\end{table*}

\begin{table}[!t]
\centering
\caption{Cumulative success rates (\%) on R1--R3 in the easy-setting.}
\label{tab:real}
\scriptsize
\setlength{\tabcolsep}{1.8pt}
\renewcommand{\arraystretch}{1.04}
\begin{tabular*}{\columnwidth}{@{\extracolsep{\fill}}llcccccc@{}}
\toprule
Task & Method & 1 & 2 & 3 & 4 & 5 & 6 \\
\midrule
\multirow{4}{*}{R1}
  & Baseline      & 70  & 60  & 60  & 60  & 50  & 40 \\
  & Ours w/o IAMoE     & 90  & 90  & 90  & 90  & 40  & 40 \\
  & Ours w/o VSR       & 100 & 90  & 70  & 70  & 70  & 60 \\
  & \textbf{Ours} & 100 & 100 & 100 & 100 & 100 & \textbf{90} \\
\midrule
\multirow{4}{*}{R2}
  & Baseline      & 100 & 80  & 80  & 80  & 60  & 60 \\
  & Ours w/o IAMoE     & 100 & 100 & 100 & 100 & 80  & 80 \\
  & Ours w/o VSR       & 100 & 100 & 100 & 100 & 90  & 90 \\
  & \textbf{Ours} & 100 & 100 & 100 & 100 & 100 & \textbf{100} \\
\midrule
\multirow{4}{*}{R3}
  & Baseline      & 100 & 70  & 70  & 70  & 40  & 30 \\
  & Ours w/o IAMoE     & 100 & 90  & 90  & 90  & 50  & 50 \\
  & Ours w/o VSR       & 100 & 80  & 80  & 80  & 60  & 60 \\
  & \textbf{Ours} & 100 & 100 & 100 & 100 & 90 & \textbf{80} \\
\bottomrule
\end{tabular*}
\end{table}

\vspace{-2pt}

\begin{figure}[!t]
\centering
\includegraphics[width=0.9\columnwidth]{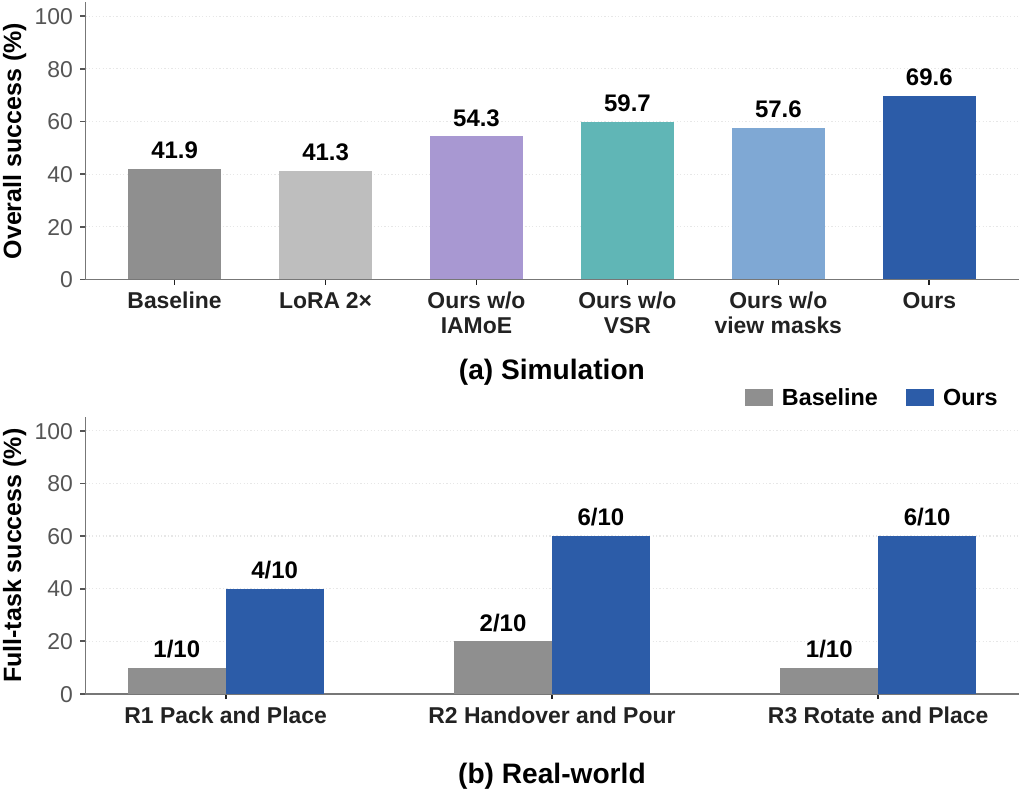}
\vspace{-3pt}
\caption{\textbf{Overall performance and robustness.}
(a) Averaged simulation success rates (\%) of matched \(\pi_{0.5}\) variants.
(b) Real-world full-task success under the Hard setting with unseen object instances and distractors.}
\label{fig:overall_results}
\vspace{-2pt}
\end{figure}

\subsection{Overall Performance against Prior Methods}

Table~\ref{tab:sim} compares the proposed framework with representative prior methods and the matched monolithic \(\pi_{0.5}\) baseline. Ours achieves the highest average success rate under both Easy and Hard settings, reaching \(81.2\%\) and \(58.0\%\), respectively, compared with \(61.5\%\) and \(22.3\%\) for the matched baseline. This corresponds to gains of \(19.7\%\) and \(35.7\%\). The improvement also extends beyond the matched baseline. Ours outperforms all prior methods in average success under both settings, while maintaining strong performance across tasks with different perceptual and interaction characteristics.

The performance gains also transfer to long-horizon real-world manipulation. As shown in Table~\ref{tab:real}, Ours achieves full-task success rates of \(90\%\), \(100\%\), and \(80\%\) on R1--R3, respectively, compared with \(40\%\), \(60\%\), and \(30\%\) for the baseline. These results show that the benefit of the proposed framework is not limited to the simulation benchmark, but extends to physical tasks requiring repeated changes in relevant visual information and bimanual interaction structure.

\subsection{Effect of Dual-Level Structural Decomposition}

To identify which aspects account for the performance gains, we analyze the single-module and structural ablations. The single-module variants confirm the distinct roles of the two decomposition levels. On S1 and S2, where perceptual heterogeneity is pronounced, \emph{Ours w/o IAMoE} consistently improves over the monolithic baseline. The effect is clear on S2, where \emph{Ours w/o IAMoE} improves over the monolithic baseline by \(23\%\) in Easy and \(37\%\) in Hard. Conversely, on S3 and S4, which emphasize interaction heterogeneity, \emph{Ours w/o VSR} improves Hard-setting success by \(28\%\) and \(30\%\), respectively. These results support the intended roles of VSR and IAMoE in addressing task-dependent visual relevance and changing inter-arm interaction structure.

The two decomposition levels also provide complementary benefits. On S5 and S6, where perceptual and interaction heterogeneity occur jointly, the full model outperforms both single-module variants under both Easy and Hard settings. Across all six tasks and settings, \emph{Ours w/o IAMoE} and \emph{Ours w/o VSR} achieve success rates of \(54.3\%\) and \(59.7\%\), respectively, whereas the full model reaches \(69.6\%\), as summarized in Fig.~\ref{fig:overall_results}(a). This indicates that selectively addressing either visual relevance or interaction structure alone is insufficient to obtain the full benefit of the proposed decomposition.

The same complementary pattern is observed in the real-world experiments. Table~\ref{tab:real} shows that Ours achieves the highest full-task success rate on all three tasks. The stage-wise results further reveal the distinct contributions of the two modules. In R1, \emph{Ours w/o IAMoE} remains effective during sequential object placement but degrades when coordinated box manipulation begins, whereas \emph{Ours w/o VSR} accumulates failures earlier as the relevant wrist view changes across stages. The full model mitigates both failure patterns, maintaining \(100\%\) cumulative success through Stage~5.

We further examine whether the improvement can be explained simply by increased fine-tuning capacity. LoRA \(2\times\), which slightly exceeds Ours in trainable parameter count while retaining the monolithic structure, achieves \(41.3\%\) overall success, essentially unchanged from the \(41.9\%\) baseline. The performance gain therefore cannot be attributed to additional LoRA capacity alone.


Finally, we evaluate the structural coupling between visual conditioning and action generation through the expert-specific view masks. Their role is most apparent during \emph{both-view + independent} segments. When only one wrist view is relevant, VSR already attenuates the opposite view, and the corresponding opposite arm is typically inactive. In contrast, when both wrist views remain relevant while the two arms act independently, VSR must preserve both views even though each arm-wise expert should directly attend only to the external and its corresponding wrist-view token positions. Without the expert-specific masks, each arm-wise expert can directly attend to the opposite wrist-view token positions, introducing unnecessary cross-arm visual conditioning and potentially encouraging task-specific correlations. Such \emph{both-view + independent} segments occur in S3--S6 but not in S1--S2. Across S3--S6, removing the view masks reduces average success from \(84.8\%\) to \(76.5\%\) in Easy and from \(56.3\%\) to \(38.0\%\) in Hard. The larger degradation under the Hard setting is consistent with the view masks restricting direct access to opposite wrist-view token positions when the two arms are generated through separate arm-wise pathways.

\subsection{Robustness under Visual Variations}
\label{sec:robustness}

Robustness is assessed by comparing performance between the Easy training distribution and the visually more challenging Hard settings. In simulation, the advantage of Ours over the monolithic baseline increases from \(19.7\%\) in Easy to \(35.7\%\) in Hard (Table~\ref{tab:sim}). Although all methods degrade under the harder visual conditions, the proposed framework retains higher performance, suggesting that explicit visual selection and interaction-aware action generation help reduce the ambiguity introduced by visual variation.

The same trend appears in the real-world Hard setting, which introduces unseen object instances and four distractors in addition to the position randomization used during training. As shown in Fig.~\ref{fig:overall_results}(b), Baseline achieves \(10\%\), \(20\%\), and \(10\%\) full-task success on R1--R3, whereas Ours reaches \(40\%\), \(60\%\), and \(60\%\). The corresponding gains of \(30\%\), \(40\%\), and \(50\%\) indicate that the proposed decomposition remains effective under unseen visual conditions.

\begin{table}[t]
\centering
\caption{Label propagation sensitivity to the manual-label fraction.}
\label{tab:label_sensitivity}
\scriptsize
\setlength{\tabcolsep}{1.4pt}
\renewcommand{\arraystretch}{0.94}

\begin{tabular*}{\columnwidth}{@{\extracolsep{\fill}}lccccc@{}}
\toprule
& & \multicolumn{2}{c}{View label} & \multicolumn{2}{c}{Interaction label} \\
\cmidrule(lr){3-4} \cmidrule(l){5-6}
Dataset & Fraction (\(N_{\mathrm{src}}\)) & Acc. & Macro-F1 & Acc. & Macro-F1 \\
\midrule
\multirow{4}{*}{RoboTwin}
& 2\% (1)   & 0.9605 & 0.9529 & 0.9788 & 0.9746 \\
& 5\% (3)   & 0.9897 & 0.9895 & 0.9928 & 0.9905 \\
& 10\% (5)  & 0.9973 & 0.9972 & 0.9981 & 0.9973 \\
& 20\% (10) & 0.9975 & 0.9974 & 0.9982 & 0.9975 \\
\midrule
\multirow{4}{*}{Real}
& 2\% (1)   & 0.7729 & 0.7542 & 0.8823 & 0.8813 \\
& 5\% (2)   & 0.8950 & 0.8879 & 0.9461 & 0.9458 \\
& 10\% (4)  & 0.9712 & 0.9675 & 0.9848 & 0.9847 \\
& 20\% (8)  & 0.9746 & 0.9713 & 0.9873 & 0.9871 \\
\midrule
\multirow{4}{*}{AIST}
& 2\% (1)   & 0.9406 & 0.9312 & 0.9262 & 0.9178 \\
& 5\% (3)   & 0.9780 & 0.9724 & 0.9561 & 0.9506 \\
& 10\% (5)  & 0.9804 & 0.9756 & 0.9641 & 0.9595 \\
& 20\% (10) & 0.9808 & 0.9762 & 0.9689 & 0.9646 \\
\bottomrule
\end{tabular*}
\end{table}

\begin{table}[t]
\centering
\caption{Router prediction agreement with manual annotations on Hard-setting rollouts.
\(N_{\mathrm{ep}}\) and \(N_{\mathrm{decision}}\) denote the numbers of rollouts and annotated router decisions.}
\label{tab:router_agreement}
\scriptsize
\setlength{\tabcolsep}{0pt}
\renewcommand{\arraystretch}{0.94}

\begin{tabular}{@{}
>{\raggedright\arraybackslash}p{0.16\columnwidth}
>{\centering\arraybackslash}p{0.10\columnwidth}
>{\centering\arraybackslash}p{0.14\columnwidth}
*{4}{>{\centering\arraybackslash}p{0.15\columnwidth}}
@{}}
\toprule
& & & \multicolumn{2}{c}{VSR} & \multicolumn{2}{c}{Action Router} \\
\cmidrule(lr){4-5} \cmidrule(l){6-7}
Outcome & \(N_{\mathrm{ep}}\) & \(N_{\mathrm{decision}}\) &
Acc. & Macro-F1 & Acc. & Macro-F1 \\
\midrule
Success & 348 & 3,123 & 0.980 & 0.973 & 0.998 & 0.998 \\
Failure & 252 & 5,112 & 0.809 & 0.776 & 0.968 & 0.943 \\
\bottomrule
\end{tabular}
\end{table}

\subsection{Router Supervision and Reliability}
We finally analyze the supervision cost and reliability of the two routers. Table~\ref{tab:label_sensitivity} evaluates the two-pass KNN propagation procedure as the manually annotated reference fraction is varied. On both RoboTwin and the real-world demonstrations, label quality improves rapidly with the reference fraction and largely saturates by \(10\%\), with only marginal gains at \(20\%\). At the \(10\%\) reference fraction, manual annotation requires on average \(2.82\) and \(12.33\) min/task for RoboTwin and the real-world data, respectively.

To examine whether this behavior is specific to demonstrations collected for our experiments, we additionally evaluate the propagation procedure on AIST-Bimanual~\cite{aist2025bimanip}, an external teleoperation dataset not collected or curated for our routing-label pipeline. AIST-Bimanual exhibits a similar saturation trend, requiring \(8.67\) min/task for manual annotation at the \(10\%\) reference fraction. These results indicate that high-quality router supervision can be obtained from a small task-specific reference set across both our demonstrations and an external data source.

Table~\ref{tab:router_agreement} evaluates router predictions against manual annotations on held-out Hard-setting closed-loop rollouts. We use the Hard setting to test whether the routers capture task-relevant routing behavior beyond correlations observed in the Easy training distribution. On successful rollouts, both routers maintain high agreement with manual annotations, indicating that their routing behavior generalizes to visually shifted environments rather than simply reproducing training-distribution patterns. Agreement decreases on failed rollouts, with a substantially larger drop for VSR than for the Action Router. Qualitative inspection of the rollout videos and router outputs shows that many VSR disagreements occur when distractor objects are confused with the task-relevant objects under the Hard setting. This suggests that view-relevance prediction is particularly sensitive to visual ambiguity introduced by distractors. In contrast, the Action Router remains comparatively stable. This difference may reflect the prediction targets. The Action Router selects a single categorical interaction mode, whereas VSR independently predicts the relevance of the left and right wrist views, allowing multiple relevance combinations. The results indicate that both routers generalize beyond the Easy training distribution, while VSR remains more sensitive to distractor-induced visual ambiguity under failed Hard-setting rollouts.



\section{Conclusion}

We presented Dual-Level Structural Decomposition for bimanual VLA policies, jointly adapting visual conditioning and action generation to changes in view relevance and inter-arm interaction structure. Simulation and real-world experiments demonstrate consistent improvements over monolithic and single-module variants, with stronger gains under challenging visual variations. These results support explicit perceptual and action-level specialization as an effective inductive bias for bimanual manipulation. The framework remains task-specific, requiring a small labeled reference set for router supervision, and uses a binary independent/coordinated interaction mode. Future work will explore task-general routing, finer-grained interaction modeling, and broader generalization across visual conditions.

{\scriptsize 
\bibliographystyle{IEEEtran}
\bibliography{references} 
}
\end{document}